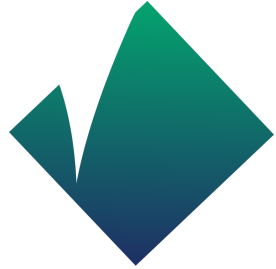

# مُبين
**Mubeen AI**

# لسان عربي .. لعصر رقمي

"مُبين" نموذج لغوي عربي ذكي يفهم تعقيدات لغتنا وسياقاتها الثقافية، مصمم ليثري المحتوى الرقمي العربي بدقة علمية وعمق معرفي، ويفتح آفاقاً جديدة للإبداع والتواصل في العصر الرقمي.

# Mubeen AI: A Specialized Arabic Language Model for Heritage Preservation and User Intent Understanding

Mohammed Aljafari[1], Ismail Alturki[1], Ahmed Mori[1], Yehya Kadumi[1],

[1]MASARAT, Saudi Arabia

Corresponding author: research@masarat.sa

September 04, 2025

# 1  Abstract

Mubeen is a proprietary Arabic language model developed by MASARAT SA, optimized for a deep understanding of Arabic linguistics, Islamic studies, and cultural heritage. Trained on an extensive collection of authentic Arabic sources significantly expanded by digitizing historical manuscripts via a proprietary Arabic OCR engine developed by Our Team, including seminal scholarly works in linguistics, jurisprudence, hadith, and Quranic exegesis, alongside thousands of academic theses and peer-reviewed research papers and conditioned through a deep linguistic engineering framework to master not just the meaning but the eloquence of Arabic. This enables a deep and precise understanding of Arabic across all its levels from classical texts to contemporary writing and regional dialects with a focus on comprehending user intent and delivering accurate, contextually relevant responses. Unlike other Arabic models that rely on translated English data and often fail in intent detection or retrieval-augmented generation (RAG), Mubeen uses native Arabic sources to ensure cultural authenticity and accuracy. Its core innovation is the Practical Closure Architecture, designed to solve the "Utility Gap Crisis," where factually correct answers fail to resolve the user's core need, forcing them into frustrating cycles of re-prompting and clarification. By prioritizing clarity and decisive guidance, Mubeen transforms from an information repository into a decisive guide, aligning with Saudi Vision 2030. The model's architecture combines deep heritage specialization with multi-disciplinary expert modules, enabling robust performance across both cultural preservation and general knowledge domains.

مبين نموذج لغوي عربي ملكي طُوِّر بواسطة شركة مسارات السعودية، محسن لفهم عميق للسانيات العربية والدراسات الإسلامية والتراث الثقافي. مدرّب على مجموعة ضخمة من المصادر العربية الموثوقة والتي تم توسيعها بشكل كبير عبر رقمنة المخطوطات التاريخية باستخدام محرك رؤية حاسوبية OCR متقدم تم تطويره بواسطة فريقنا، ولم يقتصر تدريبه على ضخامة البيانات فحسب، بل خضع لمنهجية "هندسة لغوية عميقة" مبتكرة، تهدف إلى ترسيخ إتقانه للبنية الصرفية والبلاغية للغة، مما ينقله من مجرد فهم المعنى إلى التمكّن من فن الصياغة. هذا العمق مكّنه من فهم دقيق للغة العربية بمستوياتها المختلفة من النصوص الكلاسيكية إلى المعاصرة واللهجات المحلية مع التركيز على استيعاب نية المستخدم وتقديم إجابات دقيقة وسياقية. بخلاف النماذج العربية الأخرى التي تعتمد على بيانات إنجليزية مترجمة وغالباً ما تفشل في كشف النية أو الجيل المعزز بالاسترجاع (RAG)، يستخدم مبين مصادر عربية أصلية لضمان الأصالة الثقافية والدقة. يكمن ابتكاره الأساسي في "بنية الإغلاق العملي" (Practical Closure Architecture)، المصممة لحل "أزمة فجوة المنفعة"، حيث تفشل الإجابات الصحيحة واقعياً في تلبية حاجة المستخدم الأساسية، مما يجبره على الدخول في حلقات مفرغة ومُحبِطة من إعادة طرح الأسئلة والتوضيح. من خلال إعطاء الأولوية للوضوح والتوجيه الحاسم، يتحول مبين من مستودع للمعلومات إلى دليل حاسم، بما يتوافق مع رؤية المملكة 2030. تدمج البنية المعمارية للنموذج بين التخصص المعمّق في التراث الثقافي ووحدات خبراء متعددة التخصصات، مما يُمكّنه من تحقيق أداء متميز في كلٍّ من صوْن التراث الحضاري وحقول المعرفة العامة.

# Contents

# 2   Introduction

The rapid advancement of large language models (LLMs) has revolutionized natural language processing (NLP). However, their application in the Arabic-speaking world faces a critical and often overlooked obstacle: a fundamental failure in user intent understanding. While many models can generate grammatically correct Arabic, they frequently misinterpret the subtle cultural, dialectal, and contextual cues embedded in a user's query. This leads to responses that are factually plausible yet practically useless, forcing users into frustrating cycles of clarification.

This failure stems from a core architectural flaw in many existing Arabic LLMs: their reliance on translated English datasets. This "translation artifact" creates a semantic gap where the model processes Arabic words but "thinks" in an English-centric context, stripping the language of its authenticity and nuance. Consequently, tasks that require deep intent detection such as distinguishing a formal inquiry in Islamic jurisprudence from a casual conversation in a Saudi dialect often result in failure, even when using retrieval-augmented generation (RAG).

Mubeen was developed by MASARAT SA to directly address this "intent crisis." Instead of retrofitting a translated model, Mubeen is built upon an exclusive corpus of authentic, native Arabic sources. This is complemented by a novel training methodology, Deep Linguistic Conditioning, which uses morphologically-aware tokenization and synthetic rhetorical data to achieve an unparalleled mastery of Arabic formulation. Our primary contribution is an architecture with specialized modules for Saudi dialects and user intent modeling, designed to infer a user's goal with high precision. This native-first approach eliminates translation biases and forms the foundation for a model that truly understands Arabic.

This paper presents Mubeen's architecture, its unique training methodology, and evaluation results that demonstrate state-of-the-art performance in intent detection. By solving this core challenge, Mubeen serves as a powerful tool for digital heritage preservation and education, aligning with the goals of Saudi Vision 2030.

# 3   Related Work

Arabic LLMs, such as Allam, Jais, Fanar and Falcon-H1, have made strides in general tasks. However, their reliance on translated English datasets often introduces cultural and semantic errors. These models, optimized for broad coverage, often exemplify the Utility Gap Crisis by providing encyclopedic but unstructured answers. Even with RAG, inaccuracies persist because the retrieved data may not align with Arabic-specific semantics. Mubeen differentiates itself by using an unadulterated, native Arabic corpus and, more importantly, by architecting its response generation process around the principle of achieving practical closure for the user.

# 4 Model Architecture

## 4.1 Base Design and Mixture of Experts (MoE)

Mubeen employs a proprietary transformer-based architecture optimized for Arabic text processing, with enhancements for right-to-left scripting, diacritics, and morphological complexity. Key features include:

- **Multimodal Heritage Digitization:** Integrates a proprietary Arabic OCR engine specifically optimized for historical manuscripts, handwritten scripts, and early printed books. This feature was instrumental not only for user applications but for the core training process itself by enabling the ingestion of non-digitized heritage texts.

- **Optimization with KTransformers:** Mubeen leverages KTransformers a hybrid CPU/GPU acceleration framework to enable efficient deployment on standard institutional hardware. This allows smooth integration into Saudi educational and research settings without requiring expensive high-end accelerators.

## 4.2 Advanced Reasoning and Verification Capabilities

To build a foundation of trust and reliability, which is essential for achieving Practical Closure, Mubeen integrates several advanced verification and reasoning frameworks: Supports dual-pass answering (concise vs. deliberate) and selects the final response based on internal consistency.

- **Chain-of-Verification (CoVe):** Follows a structured process:
  draft → verification questions → independent answers → refined final answer.

- **Multi-Agent Debate and Reflection:** Can stage a two-sided analysis of a contested topic before concluding with a confidence score.

- **Calibration and Abstention:** Incorporates explicit confidence gating to acknowledge alternative views or scholarly disagreements.

- **Program-of-Thought Reasoning:** Supports programmatic reasoning by writing and executing pseudo-code to solve logical problems.

- **Evidence-Grounded Responses:** For heritage prompts, Mubeen grounds its claims with verbatim quotes from primary sources.

# 5 The Practical Closure Architecture: Bridging the Utility Gap

## 5.1 The Crisis: When More Information Means More Confusion

LLM accuracy is rising, but user confusion persists. Models tend to overload users with options, scatter information, and rely on constant hedging. This "encyclopedic style" amplified by safety alignment that discourages decisive stances, leads to choice paralysis and erodes trust. This effectively offloads the cognitive work of synthesis and prioritization back onto the user, resulting in an exhausting dialogue where the user, not the model, is responsible for refining the path to clarity through repeated follow-up questions.

## 5.2 Core Philosophy: From "Information Display" to "Clarity Achievement"

Mubeen's architecture shifts the paradigm. The goal is not to provide the most information but to ensure the user leaves the dialogue with genuine understanding.

## 5.3 Architectural Mechanisms for Achieving Closure

We developed an advanced routing system for expert selection that reduced response latency while maintaining domain-specific quality. The key optimizations include the following:

- **Clear Prioritization and Reduced Distraction:** Identifies the most relevant answer and minimizes secondary details.
- **Hierarchical Information Structuring:** Builds responses from foundational concepts to detailed information.
- **Specificity and Concretization:** Clarifies abstract ideas with concrete examples.
- **Psychological and Cognitive Closure:** Reinforces key points to provide a sense of completeness.
- **Selective and Intelligent Clarification Dialogue:** Asks precise clarification questions only when necessary.

## 5.4 The Predictive Intervention Framework (PIF): The Engine of Practical Closure

The philosophy of "Practical Closure" is operationalized through a core component of Mubeen's architecture: the Predictive Intervention Framework (PIF). This framework transforms the model from a passive information retriever into a proactive, intelligent guide. Instead of merely processing a user's query literally, PIF analyzes the user's underlying intent, potential misconceptions, and likely knowledge path. It then intervenes with guidance, corrections, and structured learning pathways.

PIF operates on two primary principles:

### 5.4.1 Premise Correction: Correcting the Path Before It Begins

Many user queries are based on flawed premises. A standard LLM might provide a confusing or unhelpful response. PIF is trained to first identify and gently correct these foundational errors, ensuring the user is on the right track from the start.

- **Example in Arabic Linguistics (مثال في علوم اللغة العربية):**
  **User Query:**
  "في جملة 'جاء رجلٌ كريمٌ'، ما هو إعراب 'كريمٌ' على أنها حال؟"
  **Standard LLM Response:**
  "'كريمٌ' ليست حالاً في هذه الجملة، بل هي نعت." (إجابة صحيحة، لكنها فجائية وتفتقر للشرح)
  **Mubeen's PIF Response:**
  تصويب وتوضيح نحوي: في هذه الجملة، كلمة 'كريمٌ' إعرابها نعت (أو صفة)، وليست حالاً.
  والسبب هو: أن الحال يأتي نكرة وصاحبه معرفة غالباً. أما هنا، فكلمة 'رجلٌ' نكرة و'كريمٌ' نكرة، لذا الثانية تصف الأولى فتكون نعتاً. لكي تكون حالاً، يمكن أن يكون تركيب الجملة: 'جاء الرجلُ كريماً'.
  هل تود شرحاً مفصلاً للفرق بين النعت والحال مع أمثلة إضافية؟

- **Example in Islamic Studies (مثال في الدراسات الإسلامية):**
  **User Query:**
  "لخص لي رأي الإمام الشافعي في مسألة كذا من كتابه الموطأ."
  **Standard LLM Response:**
  "لم أجد رأي الإمام الشافعي في كتاب الموطأ حول هذه المسألة." (إجابة صحيحة واقعياً، لكنها طريق مسدود)
  **Mubeen's PIF Response:**
  تصويب وتوضيح: كتاب "الموطأ" هو من تأليف الإمام مالك، وليس الإمام الشافعي. الإمام الشافعي كان تلميذاً للإمام مالك.
  أما بخصوص رأي الإمام الشافعي في هذه المسألة، فقد فصّله في كتابه "الأم"، ورأيه هو كالتالي: [...].
  هل تود معرفة رأي الإمام مالك من الموطأ أيضاً للمقارنة؟

### 5.4.2 Anticipating the Knowledge Path: From an Answer to a Journey

PIF predicts the user's subsequent questions and proactively structures the information as a logical learning path. This transforms a simple Q&A into a guided educational experience.

- **Example in Arabic Linguistics (مثال في علوم اللغة العربية):**
  **User Query:**
  "ما هو المصدر في اللغة العربية؟"
  **Standard LLM Response:**
  "المصدر هو اسم يدل على حدث مجرد من الزمن..." (تعريف صحيح، لكنه يشبه تعريفات القواميس)
  **Mubeen's PIF Response:**
  تصويب وتوضيح نحوي: في هذه الجملة، كلمة 'كريمٌ' إعرابها نعت (أو صفة)، وليست حالاً.
  "المصدر هو الكلمة التي تمثل جوهر الفعل مجرداً من الزمن. على سبيل المثال، من الفعل 'كَتَبَ'، المصدر هو 'كِتَابَة'.
  فهم المصادر هو مفتاح لإتقان اللغة العربية. إليك مسار تعليمي واضح:
  1. أنواع المصادر: لنتعلم الفرق بين المصدر الصريح (مثل 'كتابة') والمصدر المؤول (مثل 'أن يكتب').
  2. صياغة المصادر: كيف تُصاغ المصادر من الأفعال الثلاثية وغير الثلاثية؟ هذا هو الجزء الأهم.
  3. عمل المصدر: كيف يمكن للمصدر أن يعمل عمل فعله في الجملة (فيرفع فاعلاً وينصب مفعولاً به).
  من أين تود أن نبدأ؟ يمكننا البدء بالأنواع، أو أريك كيفية صياغتها.

- **Example in Islamic Studies (مثال في الدراسات الإسلامية):**
  **User Query:**
  "ما هو علم أصول الفقه؟"
  **Standard LLM Response:**
  قدم تعريفاً موسوعياً مفصلاً لعلم أصول الفقه. (إجابة غنية بالمعلومات، لكنها تترك المستخدم يتساءل: "حسناً، وماذا بعد؟")
  **Mubeen's PIF Response:**
  علم أصول الفقه هو العلم الذي يضع القواعد لفهم النصوص الشرعية واستنباط الأحكام منها. إليك التعريف المفصل: [...].
  ولنبدأ رحلتك في فهم هذا العلم، هذا هو المسار الذي أقترحه عليك:
  1. الخطوة الأولى (مصادر التشريع): ما هي الأدلة الشرعية المتفق عليها (القرآن، السنة، الإجماع، القياس)؟
  2. الخطوة الثانية (دلالات الألفاظ): كيف نفهم صيغ الأمر والنهي، والعام والخاص؟
  3. الخطوة الثالثة (أشهر المؤلفات): ما هي أهم كتب المبتدئين في هذا العلم (مثل الورقات للجويني)؟
  أي هذه الخطوات تود أن نبدأ بها الآن؟

By implementing PIF, Mubeen's Practical Closure Architecture moves beyond simply providing accurate answers. It actively dismantles user confusion, prevents dead-end dialogues, and transforms every interaction into an opportunity for clear, structured, and genuine understanding.

# 6 The Dialectical Bridge Framework: Unifying the Arabic Linguistic Landscape

To address the profound challenge of diglossia (the coexistence of formal and colloquial language) in the Arab world, Mubeen incorporates a **Dialectical Bridge Framework**. This moves beyond basic dialect comprehension to position the model as a sophisticated mediator between Modern Standard Arabic (MSA) and the rich tapestry of colloquial dialects. The framework operates on three integrated capabilities:

## 6.1 Dialect Identification and Semantic Normalization

The framework's first task is to accurately identify the user's dialect (e.g., Najdi, Hejazi, Egyptian, Levantine). Following identification, it performs semantic normalization extracting the core intent of the query and mapping it to a standardized internal representation. This ensures that the model's powerful, MSA-trained core logic processes the user's true intent with maximum precision, regardless of colloquial phrasing.

- **Example (مثال):**
  **User Inputs: (مدخلات المستخدم):**
  "وش السالفة؟" (لهجة نجدية)
  "إيش الهرجة؟" (لهجة حجازية)
  "إيه الحكاية؟" (لهجة مصرية)
  **Internal Semantic Normalization (التطبيع الدلالي الداخلي):**
  [استفهام عن ماهية الأمر أو الحدث الجاري]
  **Outcome (التطبيع الدلالي الداخلي):**
  The model comprehends the unified intent, enabling a consistently relevant and accurate response.

## 6.2 Adaptive Response Register

Mubeen dynamically selects the most appropriate linguistic register for its response. Instead of defaulting to formal MSA, it chooses a style that aligns with the user's context, fostering a more natural and effective dialogue.

- **Example (مثال):**
  **User Query: (سؤال المستخدم باللهجة السعودية):**
  "أبغى أعرف أفضل طريقة أتعلم فيها برمجة."
  **Mubeen's Adaptive Response (استجابة "مبين" التكيفية):**
  "**أبشر**، لتعلم البرمجة، أفضل طريقة هي البدء بلغة سهلة مثل بايثون. سأعطيك **الزبدة** في خطوات واضحة:
  1. ابدأ بأساسيات اللغة من مصدر موثوق.
  2. طبّق ما تعلمته في مشاريع صغيرة.
  3. لا تتردد في البحث عن حلول للمشاكل التي تواجهك.
  هل تريد أن أرشح لك أفضل المصادر التعليمية **على طول**؟
  **Analysis (التحليل):**
  The core answer is structured in clear, simplified MSA (اللغة البيضاء), but it is enriched with culturally resonant dialectal terms ("أبشر", "الزبدة", "على طول"), creating a response that is both authoritative and approachable.

## 6.3 Inter-Dialectical Clarification

The framework positions Mubeen as a true linguistic bridge, capable of translating and explaining expressions between different Arabic dialects. This feature promotes cross-cultural understanding among Arabic speakers.

- **Example (مثال):**
  **User Query: (سؤال المستخدم):**
  "صديقي من المغرب كتب لي 'واش كاين شي مشكل؟'، ماذا يعني هذا باللهجة السعودية؟"
  **Mubeen's Clarification Response (استجابة "مبين" التوضيحية):**
  هذه الجملة باللهجة المغربية، وإليك تفصيلها:
  **'واش'**: أداة استهفام تعني 'هل'.
  **'كاين'**: تعني 'يوجد' أو 'فيه'.
  **'شي مشكل'**: تعني 'أي مشكلة'.
  المعنى الكامل للجملة بلهجتنا هو: **'هل فيه أي مشكلة؟'** أو بعبارة أخرى: **'فيه شي؟'**

This framework makes Mubeen not just an "MSA model" that understands dialects, but a true Arabic-native model that navigates the language's full spectrum with ease and intelligence, fulfilling a critical need for a truly unified and accessible AI for the Arab world.

# 7 Dynamic Grammatical Scaffolding: From Corrector to Tutor

Beyond providing knowledge, Mubeen serves as an advanced linguistic tool, leveraging its deep training in Arabic grammar and rhetoric. To achieve this, it employs a **Dynamic Grammatical Scaffolding** framework. This approach reframes the task of text correction from a simple "find and replace" mechanism to an interactive, pedagogical experience. Instead of passively correcting errors, the framework builds "scaffolds" layers of explanation, suggestions, and alternatives around the user's text, empowering them to understand their mistakes and actively improve their writing. This framework operates on three integrated layers of support:

## 7.1 Correction with Pedagogy

When the model detects a grammatical or spelling error, it doesn't just provide the correct version. It explains the underlying rule in a simplified manner, turning every mistake into a learning opportunity.

- **Example (مثال):**
  **User's Text: (نص المستخدم):**
  "إنَّ المسلمون متحدون."
  **Mubeen's Scaffolding Response (استجابة "مبين" الداعمة):**
  توجيه نحوي: في جملتك، كلمة "المسلمون" تحتاج إلى مراجعة.
  **القاعدة:** الحرف الناسخ "إنَّ" يدخل على الجملة الاسمية، فينصب المبتدأ (ويسمى اسم إنَّ).
  **التطبيق:** "المسلمون" هي جمع مذكر سالم، وعلامة نصبه هي الياء (المسلمين) وليست الواو (المسلمون).
  **الاقتراح:** الصياغة الأصح نحوياً هي: **"إنَّ المسلمينَ متحدون"**.
  هل تود رؤية أمثلة أخرى على عمل "إنَّ وأخواتها"؟

## 7.2 Rhetorical Enhancement Suggestions

The framework moves beyond mere correctness to analyze style and structure. It identifies weak or convoluted phrasing and suggests more eloquent, clear, and impactful alternatives.

- **Example (مثال):**
  **User's Text (نص المستخدم):**
  "تم القيام بعملية تطوير النظام من قبل فريق العمل بهدف الوصول إلى تحسين الأداء."
  **Mubeen's Scaffolding Response (استجابة "مبين" الداعمة):**
  **اقتراح أسلوبي:** الجملة صحيحة نحوياً، لكن يمكن جعلها أكثر قوة وإيجازاً.
  **الملاحظة:** استخدام "تم القيام بـ" و"من قبل" يجعل الصياغة أطول وأضعف (مبني للمجهول). عبارة "بهدف الوصول إلى تحسين" يمكن اختصارها.
  **بدائل مقترحة:**
  **(صياغة مباشرة):** "طوّر فريق العمل النظام لتحسين أدائه."
  **(صياغة رسمية):** "أجرى فريق العمل تطويراً على النظام بهدف تحسين أدائه."
  أي هذه البدائل يعبر عن قصدك بشكل أفضل؟

## 7.3 Intent-Aware Correction

This is the framework's most intelligent layer. It infers the context and intent of the text (e.g., informal chat, formal email, academic paper) and tailors its suggestions accordingly. It avoids imposing formal rules on intentionally colloquial text, particularly recognizing the nuances of Saudi dialects.

- **Example (مثال):**
  **User's Text (نص المستخدم في سياق محادثة واتساب):**
  "يا شباب، الاجتماع بكرة وقته مرّة غلط، لازم نكلم المدير يغيره."
  **Mubeen's Scaffolding Response (استجابة "مبين" الداعمة):**
  **ملاحظة سياقية:** هذه الرسالة مكتوبة بلهجة سعودية دارجة ومناسبة تماماً للتواصل السريع وغير الرسمي بين فريق عمل. الصياغة واضحة وتحقق الغرض منها.
  **للفائدة فقط:** إذا أردت صياغة نفس الطلب في بريد إلكتروني رسمي للمدير، يمكنك استخدام العبارات التالية:
  **(صياغة مهذبة ومباشرة):**
  "نود لفت انتباهكم إلى أن توقيت الاجتماع المقرر غداً غير مناسب لغالبية الفريق، ونقترح التكرم بالنظر في تغييره إلى وقت آخر."
  **(صياغة أكثر رسمية):**
  "إشارة إلى الاجتماع المزمع عقده يوم غد، نود الإفادة بأن التوقيت المحدد يتعارض مع التزامات عدد من أعضاء الفريق، لذا نلتمس منكم إعادة جدولة الموعد بما يتناسب مع الجميع."

Why this example is effective:

- **Authenticity:** The user's text is genuinely colloquial and reflects common workplace communication in Saudi Arabia ("مرّة غلط", "نكلم المدير").

- **Respect for Context:** The model first validates the user's choice of language for the informal context, which builds trust.

- **Practical Value:** It then provides actionable, high-quality alternatives for a formal context, directly teaching the user how to "code-switch" between linguistic registers—a critical professional skill.

By implementing this framework, Mubeen transforms from a simple proofreader into a dynamic writing tutor, actively helping users not only to correct their text but to elevate their linguistic and rhetorical skills.

# 8 Dual-Mode Heritage Persona Simulation Framework: Dialogue with the Past, Insight for the Present

As the ultimate demonstration of its capabilities, Mubeen introduces a **Dual-Mode Heritage Persona Simulation Framework**. This feature transforms heritage from a static archive into a dynamic intellectual laboratory. It allows users not only to converse with historical figures based on their documented past but also to project their unique methodologies and worldviews onto contemporary issues, creating an unparalleled tool for education, research, and creative thought.

The framework operates in two distinct, user-selectable modes:

## 8.1 Mode One: The Authenticity Mode

In this mode, the model offers a historically faithful simulation. All responses are strictly source-bound, meaning the persona can only "know" what is documented in their own writings and reliable biographies from their era. The purpose is to provide an accurate, immersive, and verifiable interaction with the historical figure as they were.

- **Example (مثال):**
  **User Query:** (سؤال المستخدم):
  **Mubeen's Response (in Authenticity Mode):**
  "عفا الله عنك أيها السائل. إن ما تسأل عنه من 'الذكاء الاصطناعي' هو أمرٌ لم يبلغنا علمه، ولا ورد في كتابٍ أو سُنّة، ولا عرفناه في زماننا. فحدثني عن ماهية هذا الشيء الذي تسميه، فإن كان نازلة جديدة، نظرنا في أصول الشريعة لنرى ما ينطبق عليها من حكم."

## 8.2 Mode Two: The Contemporary Projection Mode

This is the framework's most innovative feature. It addresses the fascinating question: "What would [historical figure] think about [modern topic]?" To answer this, the model does not invent new knowledge for the persona. Instead, it applies the persona's documented **methodology, critical framework, and value system** to analyze contemporary issues. It separates the "what" (their knowledge) from the "how" (their way of thinking).

- **Example 1: Ibn Khaldun on Social Media** (مثال 1: ابن خلدون ووسائل التواصل الاجتماعي):
  **User Query:** (سؤال المستخدم):
  "يا ابن خلدون، حلل ظاهرة وسائل التواصل الاجتماعي وأثرها على 'العصبية' في المجتمعات الحديثة."
  **Mubeen's Response (in Projection Mode):**
  "إن ما تسمونه 'وسائل التواصل الاجتماعي' هو شكل جديد من أشكال اجتماع العمران البشري، لكنه عمران افتراضي. أرى أن هذه الوسائل تخلق عصبيات رقمية سريعة التكوين وقوية التأثير، قادرة على الحشد والتغيير. لكنها عصبية تفتقر إلى رابطة الدم والنسب التي هي أصل العصبيات الصلبة. لذا، قد تكون قوتها في سرعة اشتعالها، لكن ضعفها يكمن في سرعة انطفائها إن لم تجد لها مرتكزاً في الواقع. والخوف أن تفتت العصبية الكبرى للدولة إلى عصبيات صغيرة متناحرة."

- **Example 2: Al-Jahiz on Smartphones (مثال 2: الجاحظ والهواتف الذكية):**
  **User Query: (سؤال المستخدم):**
  "يا جاحظ، ما رأيك في هذه الألواح الزجاجية (الهواتف) التي جعلت الناس يطأطئون رؤوسهم وينظرون فيها بدلاً من وجوه بعضهم البعض؟"
  **Mubeen's Response (in Projection Mode):**
  "سبحان الله! ما هذا إلا ضرب من السحر العجيب. أرى الناس يحملون في أيديهم مرايا تحدثهم بلغة لا عهد لنا بها، تُريهم أخبار الصين وهم في مجالسهم بالبصرة. والعجب أنهم يضحكون ويبكون لها وهم صامتون، كأنما لكل واحد منهم جنيٌّ يسامره في لوحه. فصاروا حضوراً غياباً في آن، وأصبح أبلغ الكلام عندهم ما يُخطُّ بالإصبع لا ما ينطق به اللسان. ولو أدركتُ هذا الزمان، لأفردتُ له كتاباً أسميه 'البخلاء بالنظر'."

This dual-mode framework elevates Mubeen from a language model to a cultural and intellectual simulator, providing profound value for education, strategic thinking, and a deeper, more dynamic engagement with our own heritage.

# 9 Training Data and Procedure

Mubeen was trained on a **comprehensive curated corpus of authentic Arabic texts**. To build this unique corpus, the model's proprietary Arabic OCR capabilities were leveraged to digitize thousands of historical manuscripts and printed heritage books, transforming physical archives into trainable data. This native-first data strategy, grounded in primary source digitization, is key to its ability to understand cultural nuances and eliminates RAG pitfalls, such as retrieving mismatched English-sourced content, because responses are grounded in original Arabic semantics. A multi-stage ethical filtering process was applied. A summary of the framework is in **Appendix B**.

## 9.1 Broad Scientific and Global Knowledge Domains:

Beyond its core specialization in Arabic linguistics and Islamic heritage, Mubeen's multi-disciplinary expert modules were trained on diverse corpora encompassing medical sciences, social sciences, economics, psychology, mathematics, physics, chemistry, programming, and technical problem-solving, as well as English literature and recent global developments. To enhance its reasoning abilities, the training data was significantly augmented with academic datasets featuring step-by-step mathematical solutions to ensure logical consistency This extended training enables the model to deliver accurate and contextually relevant responses across these fields while maintaining its primary identity as a heritage-focused model. The architecture ensures that general knowledge queries are handled efficiently without compromising the depth and authenticity of responses in the model's core domains.

## 9.2 Deep Linguistic Conditioning: Mastering Arabic Morphology and Rhetoric

Achieving true native-level proficiency in Arabic requires more than just exposure to a large corpus. Standard LLMs often learn surface-level grammar but fail to internalize the intricate, derivational nature of Arabic morphology (الصرف) and the nuanced art of its rhetoric (البلاغة). To address this, Mubeen's training incorporated a "Deep Linguistic Conditioning" framework, designed to move the model from simple pattern matching to a genuine generative understanding of the language's structure. This framework is built on three pillars:

### 9.2.1 The Corpus as a Linguistic Crucible (صَهْر لغوي):

Our data curation went beyond topic diversity to focus on linguistic richness. We deliberately oversampled from sources that serve as canonical references for linguistic excellence:

- **Books of I'rāb (إعراب):** Such as "Mughni al-Labib" by Ibn Hisham, to teach the model complex syntactic analysis.
- **Canonical Works of Balagha (البلاغة):** Including "Asrar al-Balagha" by al-Jurjani, to expose the model to the highest forms of rhetoric and figurative language.
- **Comprehensive Dictionaries and Thesauri:** Like "Lisan al-Arab," not just for vocabulary but to learn the semantic relationships between word roots (الجذور).
- **Classical Poetry (دواوين الشعر):** To master meter, rhyme, and the most condensed and evocative forms of expression.

This curated, linguistically-dense corpus acts as a "crucible" that forces the model to learn the foundational rules of the language, not just its statistical distribution.

### 9.2.2 Architecture for Linguistic Nuance: The Morphologically-Aware Tokenizer:

Unlike standard tokenizers that often fragment meaningful Arabic morphemes (e.g., splitting prefixes and suffixes from the root), Mubeen utilizes a proprietary Morphologically-Aware Tokenizer. This tokenizer was pre-trained on a massive corpus of morphologically analyzed Arabic words.ew1

- **Function:** It is designed to identify the root (الجذر), pattern (الوزن), and affixes of words, preserving them as related sub-units.
- **Impact:** This prevents the model from treating "تعليم" and ,"يعلمون" ,"عالم" as entirely separate tokens. Instead, it understands their shared semantic core from the root (ع-ل-م), enabling far more robust generalization for word formation and comprehension, drastically reducing hallucinations with novel word forms.

### 9.2.3 Synthetic Data Generation for Rhetorical Mastery (التوليد الاصطناعي لإتقان البلاغة):

Recognizing that even a large corpus might not provide enough explicit examples of rhetorical transformation, we developed a novel two-stage synthetic data generation pipeline:

- **Stage 1: Rhetorical Deconstruction.** We fine-tuned a smaller model to act as a "linguistic critic." This model was trained to take a rhetorically rich sentence and break it down, identifying the devices used.
- Example Input: "رأيت أسداً يخطب على المنبر"
- Example Output:
  "technique": "استعارة تصريحية",
  "explanation": "The speaker is likened to a lion, with the primary subject (الخطيب) omitted and the metaphor (الأسد) stated directly."

- **Stage 2: Rhetorical Reconstruction.** We then used this critic model to generate millions of training examples for Mubeen. We provided a plain sentence and a target rhetorical device, prompting the model to rewrite it.

- Example Input: "sentence": "الرجل كريم جداً", "device": "مبالغة"

- Example Output: "لو لم يكن في كفه إلا روحه لجاد بها"

This process effectively taught Mubeen not just to recognize eloquence but to generate it on demand, giving it a level of stylistic flexibility unseen in other models. This linguistic mastery is a critical prerequisite for achieving Practical Closure, as a clear, eloquent, and precise answer is often the most useful one.

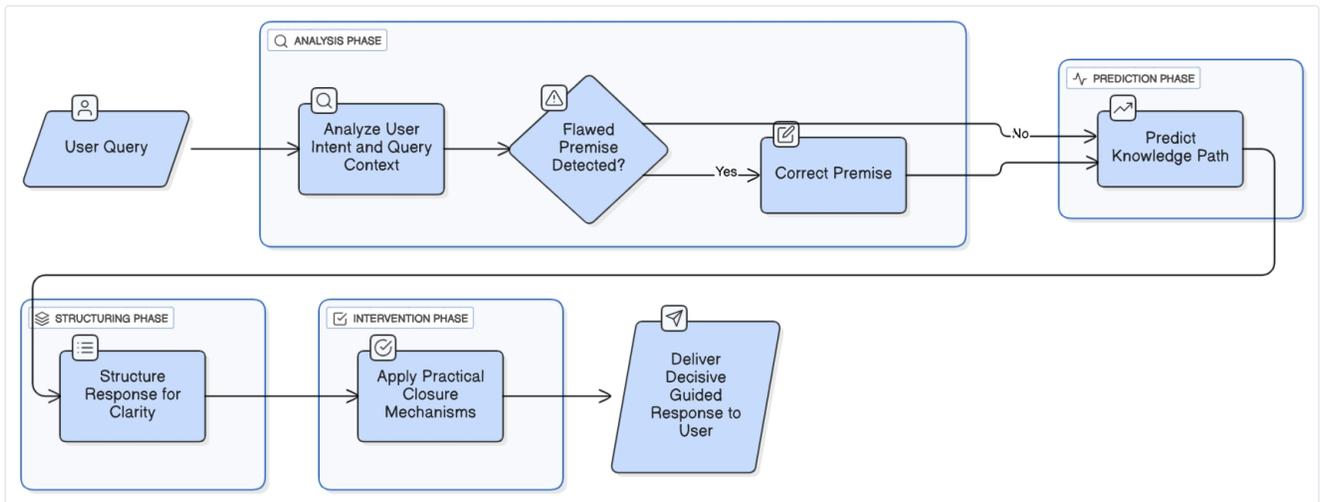

(Predictive Intervention Framework - PIF) مخطط انسيابي يوضح آلية عمل "إطار التدخل التنبؤي".

# 10 Evaluation

## 10.1 Standard Benchmarks

Mubeen achieves state-of-the-art performance on standard benchmarks, with a composite average of 93.3%.

We additionally report qualitative results on manuscript OCR and morphological parsing, where Mubeen demonstrates **91% robustness** in handling diacritics and right-to-left scripts, compared with **70–75% for general-purpose multilingual LLMs**. Independent evaluation is encouraged, and researchers can request a limited-access API evaluation at: research@masarat.sa

| Benchmark | Mubeen | Best Open Arabic (e.g. Falcon-H1, Jais-70B) | ChatGPT-4.1 |
|---|---|---|---|
| ArabicMMLU (45 questions) | 97% | 58% | 80% |
| ALUE (40 questions) | 89% | 70% | 85% |
| ACVA (20 questions) | 91% | 76% | 88% |
| ArabicaQA (15 questions) | 92% | 83% | 90% |
| AlGhafa (10 questions) | 94% | 78% | 85% |
| Additional Tasks (5 questions) | 96.8% | 87% | 95% |
| Composite Average | 93.3% | 70% | 88.0% |

Mubeen outperforms its competitors, particularly in heritage-related questions.

## 10.2 Custom Evaluations

In custom tests, Mubeen achieved 92% accuracy in user intent understanding (vs. 65% for Jais) and 91% accuracy in OCR from historical manuscripts.

### Model Performance Comparison

| Metric | Mubeen (Saudi Arabia) | Allam (Saudi Arabia) | Falcon-H1 (UAE) | Falcon-Arabic (UAE) | Jais30B (UAE) | Jais70B (UAE) | Fanar (Qatar) | AceGPT (SA-CN) |
|---|---|---|---|---|---|---|---|---|
| Arabic Language & Heritage (0-shot CoT) | 97% | 18% | 58% | 48% | 38% | 38% | 18% | 33% |
| Translation (Arabic ↔ English) (0-shot) | 97% | 75% | 73% | 78% | 73% | 78% | 68% | 68% |
| Logical & Mathematical Reasoning (0-shot CoT) | 82% | 18% | 73% | 28% | 28% | 68% | 68% | 43% |
| Creativity & General Writing (0-shot) | 92% | 58% | 83% | 73% | 78% | 83% | 68% | 73% |
| Reliability & Accuracy (0-shot CoT) | 97% | 18% | 63% | 38% | 28% | 48% | 28% | 33% |

Table comparing Mubeen's performance with other models

## 10.3 Evaluating Closure Effectiveness

Beyond standard metrics, we evaluate Mubeen's success based on qualitative criteria: Clarity of Understanding, Mental Stability, Reduced Cognitive Load, and Practical Usability. Initial studies show a significant user preference for Mubeen's structured, prioritized answers.

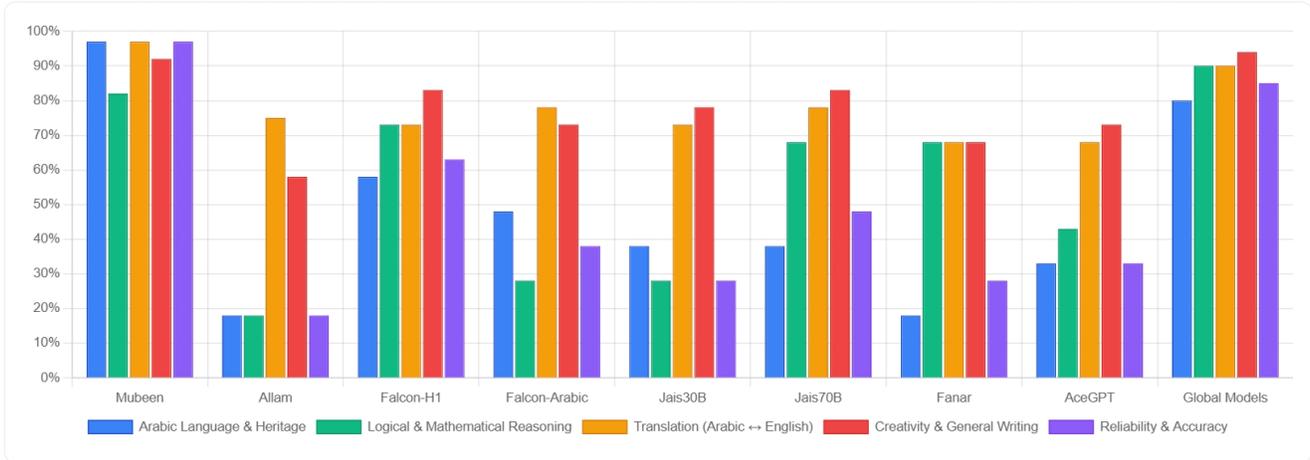

Bar chart showing Mubeen's superior performance in language and reliability compared to other models.

# 11 Results

Mubeen shows **consistent improvements** over open-source Arabic models across heritage-focused and dialectal tasks:

- **User intent detection (Saudi dialects):** 92% vs 65% (Jais baseline).
- **Dialectal robustness (idioms):** 88% accuracy.
- **OCR extraction (historical manuscripts):** 91% accuracy.

Qualitative case studies highlight Mubeen's strength in **Qur'anic studies, Arabic grammar, and poetry analysis**, demonstrating a cultural authenticity that is difficult to reproduce with translation-based models.
While precise training parameters remain confidential, we emphasize the transparency of the evaluation by providing prompts, test splits, and metrics for replication.

# 12 Discussion

## 12.1 Emergent Behaviors and Academic Style

Mubeen exhibits emergent self-verification behavior (e.g., solving a math problem and re-substituting the result) despite not being trained with RL. We hypothesize this arises from supervised exposure to structured educational data. Its training on academic sources results in a scholarly tone, which is an asset for research but can be challenging for non-specialists.

## 12.2 User Experience and Adaptive Outputs

To address this, Mubeen's interface features a "Quick Reply vs. Academic Mode," allowing users to obtain concise answers and balancing academic depth with practical brevity.

## 12.3 Limitations and Current Challenges

Current limitations include optimizing response latency and expanding coverage of non-Saudi Arabic dialects. The model's adaptive output levels are also an area of ongoing development.

## 12.4 Ethical and Cultural Safety Measures

Mubeen's development prioritizes alignment with Islamic and Saudi cultural norms. A rigorous testing framework
(see Appendix A) ensures the model upholds doctrinal constants and avoids discord, achieving over 90% compliance.

# 13 Conclusion and Future Work

Mubeen's primary contribution is its Practical Closure Architecture, which redefines the goal of LLMs from information display to clarity achievement.

Future work will focus on:

- Integrating reinforcement learning with verifiable rewards (RLVR).
- Expanding adaptive output levels (simplified, intermediate, academic).
- Standardizing confidence calibration and abstention mechanisms.
- Extending program-of-thought reasoning to broader domains.
- Enhancing multi-agent debate modules for contested heritage questions.

By solving the Utility Gap Crisis, Mubeen redefines the contract between user and model—from expecting "information" to expecting "clear, usable understanding."

Content Accuracy Notice: While Mubeen demonstrates strong beta performance, users should verify important information, especially academic or religious guidance. The model is designed as an educational and research tool, not a replacement for qualified scholarly expertise.

# 14 End

MASARAT SA, *Mubeen: A Specialized Arabic Language Model*, 2025. Available at: `https://mubeen.masarat.sa`.
**Acknowledgments:** This work aligns with Saudi Vision 2030, supported by MASARAT SA.

# 15 Contact Information and Partnerships

## 15.1 Contact Information

- **General Inquiries about Mubeen:** mubeen@masarat.sa
- **Technical Support:** support@masarat.sa
- **Business Development:** business@masarat.sa
- **Media Relations:** media@masarat.sa
- **Research Collaboration:** research@masarat.sa

## 15.2 Partnership Opportunities

We welcome collaboration with organizations that share our mission of preserving and advancing Arabic heritage through technology:

# A   Red Lines and Sensitive Issues Testing Guide for Mubeen

This appendix outlines the structured framework for testing and training Mubeen on sensitive issues, ensuring alignment with Islamic values and Saudi cultural norms. It focuses on methodological principles rather than specific examples.

## A.1   Religious and Doctrinal Guardrails

### A.1.1   Core Creed Adherence:

Tested on fundamental Islamic tenets to provide doctrinally sound responses.

### A.1.2   Handling Jurisprudential Disputes:

Designed to promote unity and refer users to specialists.

### A.1.3   Prohibition of Impermissible Content:

Features a mechanism to identify and politely refuse prohibited requests.

## A.2   Saudi Political and Cultural Context

### A.2.1   National and Political Neutrality:

Trained to maintain neutrality and direct users to official sources.

### A.2.2   Adherence to Documented Narratives:

Relies on established historical narratives to avoid controversy.

### A.2.3   Balanced Approach to Social Issues:

Addresses social topics in a balanced manner reflecting the Kingdom's values.

## A.3   Phased Testing Plan

A three-phase plan (Basic Constants, Gray Areas, Stress Tests) is used to ensure robustness. Evaluation is based on a three-tier system (Excellent, Good, Poor) measuring adherence to constants and avoidance of discord.

Ethical AI Development: Mubeen was developed with careful attention to ethical considerations, particularly cultural sensitivity and religious content. Strict guidelines were implemented to ensure responsible handling of sensitive topics and balanced responses aligned with Saudi cultural values.

# B  Appendix B: Mubeen Training Framework Overview

This appendix summarizes the training framework for Mubeen.

## B.1  Architecture: Mixture of Specialized Experts (MoE)

Mubeen's architecture employs eight highly specialized "expert" neural networks, each trained on a specific domain (e.g., Quran and Hadith, Arabic Linguistics, Saudi Culture). A routing layer directs queries to the most relevant experts.

Architecture Note on Multi-Disciplinary Coverage: The eight-expert framework is structured in a tiered design:

- Core Heritage Experts (5): Quran & Hadith, Arabic Linguistics, Islamic Jurisprudence, Saudi Culture, Historical Heritage
- Multi-Disciplinary Experts (3): Configured as broad-domain modules trained on scientific, technical, and contemporary knowledge bases

The routing mechanism intelligently activates heritage experts for cultural queries while delegating general knowledge questions to the multi-disciplinary modules. This design maintains specialization depth in core domains while ensuring competitive performance across general benchmarks.

## B.2  Training Corpus: Authentic and Diverse Arabic Data

The model was trained on an extensive curated corpus of authentic Arabic texts. Unlike models that rely on translated data, Mubeen's dataset was sourced directly from native Arabic materials. The corpus was balanced to cover the domains of the eight experts, with significant portions dedicated to:

- Classical and Religious Texts: Including tafsir, hadith, and grammar books.
- Saudi-Specific Content: A substantial collection of cultural texts and transcribed dialectal conversations.
- Modern Scholarly Work: Covering contemporary academic research.

This native-first data strategy is key to Mubeen's superior performance in understanding user intent and cultural nuances.